# Generative Artificial Intelligence: Implications for Biomedical and Health Professions Education

William Hersh, MD
Professor
Department of Medical Informatics & Clinical Epidemiology
School of Medicine
Oregon Health & Science University
Mail Code: BICC
3181 SW Sam Jackson Park Rd.
Portland, OR, USA, 97239
Email: hersh@ohsu.edu
ORCID: 0000-0002-4114-5148

Abstract

Generative AI has had a profound impact on biomedicine and health, both in professional work and in education. Based on large language models (LLMs), generative AI has been found to perform as well as humans in simulated situations taking medical board exams, answering clinical questions, solving clinical cases, applying clinical reasoning, and summarizing information.  Generative AI is also being used widely in education, performing well in academic courses and their assessments. This review summarizes the successes of LLMs and highlights some of their challenges in the context of education, most notably aspects that may undermines the acquisition of knowledge and skills for professional work. It then provides recommendations for best practices overcoming shortcomings for LLM use in education. Although there are challenges for use of generative AI in education, all students and faculty, in biomedicine and health and beyond, must have understanding and be competent in its use.

1. Introduction

The overall goal of this review is present the key issues for generative artificial intelligence (AI) in the context of biomedical and health professions education. Biomedical and health professions students are defined broadly as any individual who works professionally in healthcare, individual health, public health, and research. This includes clinicians such as physician, nurses, pharmacists, etc. as well as those who do not provide clinical care, such as administrators, project managers, researchers, educators, and more. Also among these are those who work in biomedical and health informatics, data science, and related fields. Although the review is focused on AI in biomedicine and health, it will draw on areas outside this broad domain where additional perspectives may be insightful.

AI has been defined in many ways, and this author prefers the notion that it consists of information systems and algorithms that are capable of performing tasks that are associated with human intelligence.(1) AI is sometimes classified into two broad categories:(2)
- Predictive AI – use of data and algorithms to predict some output, e.g., diagnosis, treatment recommendation, prognosis, etc.
- Generative AI – generates new output based on prompts, e.g., text, images, etc.

A large part of modern success of AI is due to machine learning (ML), whose origins have been attributed to the Arthur Samuel in the 1950s and which has been defined as "computer programs that learn without being explicitly programmed."(3) The most success applying ML to AI has come from so-called deep learning, which is based on many-layered neural networks.(4) Predictive AI has typically used supervised ML, where models are trained with labeled or annotated data split into training and test data. On the other hand, generative AI systems are mostly trained with unsupervised ML due to the much larger sizes of training data and their generative, i.e., non-predictive uses, although some supervised ML takes place to avoid inappropriate generation, e.g., racist or sexist language.(5)

Generative AI had been around for a number of years but came to the fore with the release of ChatGPT by OpenAI on November 30, 2022. Based on large language models (LLMs) trained using deep learning with massive amounts of digital text, generative AI systems have had profound impact in biomedicine and health, including in education. A number of reviews describe LLMs generally(6,7) as well as their applications in biomedicine.(8–10)

This review will focus on generative AI, with a particular perspective on its implications for biomedical and health education. Some of the discussion may go beyond education, but only if there is some relevance to education. The review will begin by describing the use of generative AI. It will then discuss the successes and limitations of generative AI, especially those that apply to education. The review will then cover issues specific to education, describing challenges and accomplishments, and then wrapping up with future directions.

2. Usage of Generative AI

Since the introduction of ChatGPT and other systems, generative AI has been rapidly taken up by a large proportion of people around the world, including biomedical and health professionals, students, consumers, and others. A survey from mid-2023 found that 56% of college students stated they had used AI on assignments or exams, with 54% agreeing that the use of AI tools on college coursework counts could, in some instances, constitute cheating or plagiarism.(11) A survey from mid-2024 of teachers, students, and parents found 49-52% each group using generative AI frequently, 18-33% using it occasionally, and less than a quarter of each group stating they had never used it.(12) Both of these surveys noted positive views of AI but also lack of consistent or comprehensive policies by educational institutions and faculty within them.

Other surveys have looked at larger populations and noted similar large-scale uptake of generative AI. One analysis from August 2024 found that 39% of working-age adults use generative AI.(13) It also found that than 24% of such individuals used it in their work at least once in the week, and about 10% used it every day at work. The authors of the survey noted that the adoption of generative AI has been faster than even adoption of personal computers or the internet. Another survey, this one of over 1000 physicians from the United Kingdom, found that 20% reported using generative AI tools in clinical practice.(14) Among those who used such tools, 29% reported using it to generate documentation after patient appointments and 28% to develop a differential diagnosis.

A survey of 2428 US adults of varying age, geographic location, and race/ethnicity carried out by the Kaiser Family Foundation found that about two-thirds of respondents reported some use or interaction with AI.(15) However, use of AI chatbots for health information and advice was much lower than for social media or Internet search (Table 1).

Table 1 – frequency of use of sources for health information and advice, adapted from (15).

| Source | Every day | At least once a week | At least once a month | Occasionally | Never |
| --- | --- | --- | --- | --- | --- |
| Social media | 46% | 10% | 4% | 18% | 22% |
| Internet search | 42% | 21% | 9% | 21% | 7% |
| AI chatbots | 6% | 7% | 4% | 19% | 63% |

3. Results of Generative AI in Biomedicine and Health

An avalanche of research has been published on the results of generative AI applied to biomedical and health tasks. Many of these studies have focused on answering questions, especially those on examinations, including medical board tests. Others have expanded to solving clinical cases or carrying out other clinical tasks, although few involve actual users and almost none have been implemented in real-world settings. Nonetheless, these results impact education in they can carry out tasks used in educational activity, including student practice and assessments. The following categories of research will be described in this section:
1. Medical Board Examinations
2. Other Medical Examinations
3. Answering Clinical Questions
4. Solving clinical cases
5. Assessing Clinical Reasoning
6. Summarization Tasks
7. Predictive Tasks
8. Performance in Graduate Courses in Biomedicine and Health
9. Performance in Academic Courses Beyond Biomedicine and Health

3.1 Medical Board Examinations

One of the most widely used data sets for LLM evaluation comes from a set of sample questions from the US Medical Licensing Exam (USMLE).(16) This multiple-choice question (MCQ) exam is taken in three steps during the second and fourth years of medical school and then in the first year out of medical school. The dataset, called MedQA, includes 12,723 questions in English (and others in Chinese). An early and highly-publicized success of the original ChatGPT was to achieve a score of over 60% on the MedQA dataset using the GPT-3.5 LLM, which would equate to a passing grade.(17) This level was surpassed by other LLMs, including GPT-4(18) and Google's Med-Gemini.(19) The most recent leader in the "USMLE arms race" is OpenAI's o1 model, scoring 96.0% (Figure 1).(20,21) LLMs have even been found to perform well on the "soft skills" portions of the USMLE exam, e.g., communication skills, ethics, empathy, and professionalism.(22)

This has led to the assessment of LLMs on other US-based medical board exams, with similar success being found, among others, in the fields of radiology,(23,24) neurosurgery,(25) and clinical informatics.(26) LLM success on board exams in not limited to the US. In Israel, the scores of all resident physicians who completed board exams in the specialties of pediatrics, internal medicine, psychiatry, obstetrics/gynecology, and general surgery in 2022 were compared to GPT-3.5 and GPT-4.(27) As seen in Figure 2, GPT-4 passed the exams in all of the specialties, scoring comparable in internal medicine and general surgery, better in psychiatry, and inferiorly in pediatrics and obstetrics/gynecology.

Figure 1 – incremental progress of LLMs on MedQA (USMLE) dataset over time.(20,21)

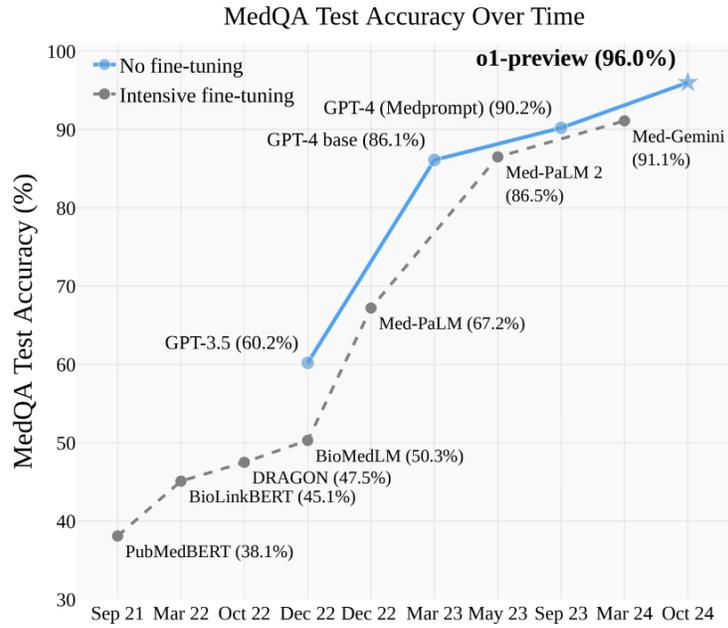

Figure 2 – LLM success versus physicians completing residency training in 5 specialties Israel. Range is shown for all test takers, human and LLM. The dotted line for each the passing level for the exam for each specialty.(27)

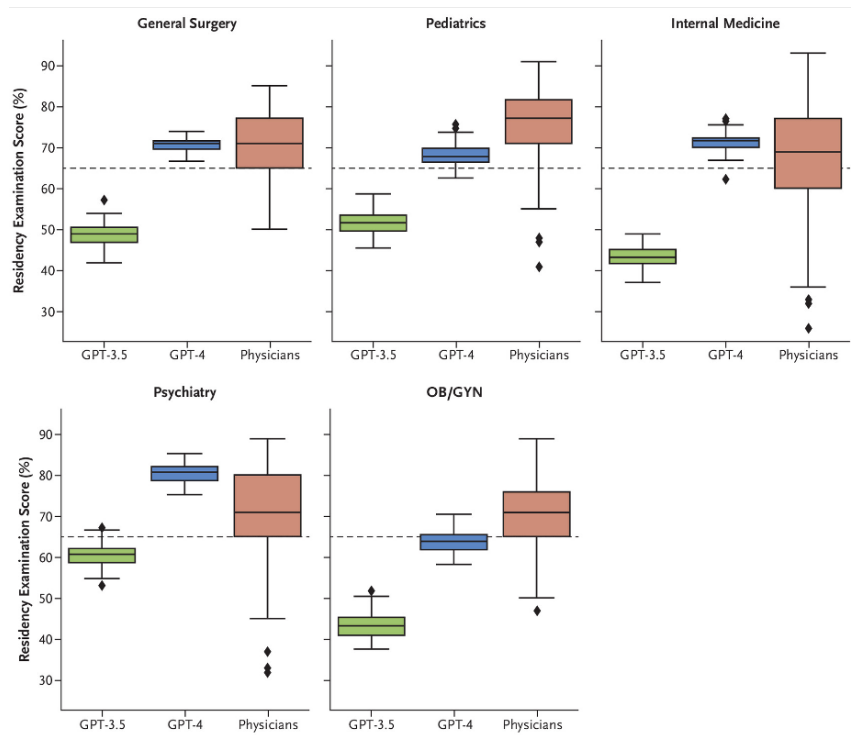

## 3.2 Other Medical Examinations

Success with generative AI has been shown with other kinds of physician examinations. On an objective structured clinical examination (OSCE), which provides a series of questions simulating a clinical encounter, ChatGPT-3.5 scored better than a group of physicians in Singapore in a simulation of assessment for membership in the Royal College of Obstetricians and Gynecologists.(28) Likewise, ChatGPT-4 achieved 81% correct on the Ophthalmology Knowledge Assessment Program (OKAP)(29) and 73.3% on the Nephrology Self-Assessment Program (nephSAP).(30).

Many medical specialties give in-training exam to residents and fellows in training. On a family medicine in-training exam, ChatGPT-4 scored 86.5% and outperformed third-year resident national averages in one study(31) and 84% on another.(32). The latter study also found that GPT-4 was able to integrate new information and carry out self-correction when needed. Likewise on a radiology in-training exam, GPT-4 achieved results comparable to between second-year and third-year radiology residents.(33) On this exam, performance on image-based questions was significantly poorer at 45.4% compared to text-only questions at 80.0%. As has been seen in some other studies, when questions were given again, GPT-4 chose a different answer about 25% of the time. Fine-tuning the LLM did not improve accuracy in this study.

## 3.3 Answering Clinical Questions

The success of generative AI is not limited to exams. Other research has assessed their ability to answer clinical questions in a variety of medical disciplines. One of the most studied applications areas has been cancer. One study assessed ChatGPT output for concordance with National Comprehensive Cancer Network treatment guidelines for breast, prostate, and lung cancer. The study found an overall concordance of 61.9% of the time, with 34.3% of outputs recommending one or more nonconcordant treatments. The study also found that responses were hallucinated, i.e., were not part of any recommended treatment, in 12.5% of outputs.(34) Another study assessed several different LLMs and found varying performance on different categories of oncology questions, with ChatGPT-4 the only LLM to score greater than 50%.(35)

An additional study assessed the ability of GPT-4 to answer clinically relevant questions regarding the management of patients with pancreatic cancer, metastatic colorectal cancer, and hepatocellular carcinoma from clinical practice guidelines.(36) Results found that the addition of information for use of retrieval-augmented generation (RAG), a method that allows new information to be added to LLMs already trained, and resulted in correct responses in 84% of cases compared to only 57% without the use of RAG.

The success of LLMs is not limited to cancer. One study of 284 questions developed by physicians found that ChatGPT-4 had highly accurate and complete answers and performed better than ChatGPT-3.5.(37) Another study looked at ChatGPT-3.5 answering

MCQs about human genetics.(38) ChatGPT-3.5 responses were found to be correct comparably to human answerers (68.2% vs. 66.6% respectively. It was also found that both ChatGPT and humans performed better on memorization-type questions than on critical thinking questions. Another interesting finding was that when asked the same question multiple times, ChatGPT provided different answers 16% of the time, including for both initially correct and incorrect answers, and gave plausible explanations for both correct and incorrect answers.

An additional study assessed GPT-4 for concordance with recommendations made by 12 physicians from a hospital consultation service.(39) There were 31 of 66 questions that had a majority (i.e., more than 6) of physicians rating concordance. Of these, the responses to 13 questions were designated concordant, 15 discordant, and 3 were unable to be assessed. The responses from GPT-4 were found to largely be devoid of overt harm, but less than 20% of the responses agreed with answer from the consultation service. It was noted that some responses contained hallucinated references.

Other studies have focused on specific types of questions for clinicians. One looked at the ability of GPT-4 to appropriately perform probabilistic reasoning by comparing performance with a large survey of human clinicians.(40) The LLM was more accurate than human clinicians in determining pretest and posttest probability after a negative test result for 5 cases but did not perform as well after positive test results. A couple of additional studies have focused on answering clinical questions using general LLMs further tuned with clinical information resources. One system called Almanac, an LLM framework using RAG from curated medical resources for medical guideline and treatment recommendations, was found to obtain significant improvement in performance compared with standard LLMs on measures of factuality, completeness, user preference, and adversarial safety.(41) Another system based on RAG and called ChatRWD outperformed the commercial system OpenEvidence as well as several general LLMs on questions seeking additional clinical evidence.(42)

There are many other medical question-answering datasets that have been developed and used to compare generative AI systems. One recent analysis used a large number of them to compare the new OpenAI o1 LLM that makes use of a new type of prompting called chain of thought prompting, where prompts to the LLM describe the task iteratively, and found that it surpassed GPT-4 in accuracy by an average of 6.2% and 6.6% respectively across 19 datasets and two newly created complex question-answering scenarios.(43) However, the study also noted several weaknesses of the LLMs, including hallucination, inconsistent multilingual ability, and discrepant metrics for evaluation. We can see from all of these studies that generative AI is very good at answering questions on medical examinations, yet is far from perfect and does suffer from incorrect and/or inconsistent answers.

3.4 Solving Clinical Cases

Some analyses of generative AI have gone beyond answering questions to assessing the ability to solve clinical cases. One line of research has involved a collection of clinical vignettes that were originally developed in the mid-2010s to assess the performance of clinical symptom-checkers.(44) Most systems using them performed poorly at the time, and certainly far worse than physicians, who averaged about 72% accuracy. One study assessed ChatGPT-3.5 with this dataset for first-pass diagnostic and triage decision accuracy, finding that ChatGPT-3.5 identified illnesses with 75.6% first-pass diagnostic accuracy and 57.8% triage accuracy.(45) This study also found that ChatGPT was useful for generating new vignettes written both for those with high and low health literacy levels. Another study found that ChatGPT-3.5 identified the correct diagnosis in the top three diagnoses for 88% of cases, compared to 54% for lay individuals and 96% for physicians.(46) ChatGPT-3.5 was also found to triage 71% of cases correctly, similar to lay individuals, with both worse than physicians, who triaged correctly for 91% of the cases.

Other studies have used the well-known clinicopathologic conferences (CPCs) featured in the New England Journal of Medicine (NEJM). These cases are deemed to be challenging and often stump even Harvard Medical School faculty. One study found that GPT-4 provided the correct diagnosis within the differential diagnosis in 64% of the 70 cases assessed and as the top diagnosis in 39% of cases.(47) Another study found GPT-4 to be correct for 57% of 38 cases assessed, which was better than almost all online readers who answered the challenge.(48) A follow-on analysis compared performance for GPT-4 with cases that were newer and older than the September 2021 training date of GPT-4, aiming to control for possible data leakage, i.e., the early cases potentially being in the training data. This analysis found comparable results in newer and older cases.

Another clinical problem-solving study assessed Google's Med-PaLM 2, which had been optimized for diagnostic clinical reasoning.(49) Unlike earlier LLM studies, this one employed real physicians solving 302 NEJM clinicopathologic conferences. Generalist physicians were given versions of cases that were redacted for diagnostic testing and the final diagnosis. The physicians were asked to generate a differential diagnosis when randomized to one of two conditions, one having access to standard online search tools or being provided output from Med-PaLM 2. Specialist physicians were provided access to the gold standard-evaluated differential diagnosis lists and asked to evaluate the lists for inclusion of the final diagnosis, comprehensiveness of the differential diagnosis, and appropriateness of the differential diagnosis. The overall best diagnosis in the top 10 of the differential diagnosis came from Med-PALM 2 only (59.1%), followed respectively by generalist physicians with Med-PALM 2 (51.7%), generalist physicians with access to conventional search systems (44.4%), and unassisted generalist physicians (33.6%). Also of note, this Google LLM was found to exceed the performance of GPT-4.

An additional dataset for clinical problem-solving has been a collection of 36 clinical vignettes from the MSD Clinical Manual, formerly known as the Merck Manual.(50) In this study, ChatGPT was found to achieve overall correctness on all questions for all cases 71.7% of the time. It performed best for answering the final diagnosis and lowest for generating an initial differential diagnosis. The overall accuracy was lower for diagnostic and management questions than for diagnosis questions. There was no variation in answer correctness by the age, gender, or acuity of the patient in the vignette.

Some studies have looked at solving clinical cases in specific settings. One study from a Dutch emergency department (ED) retrospectively reviewed notes and entered them into ChatGPT-3.5 or ChatGPT-4.(51) For generating differential and leading diagnoses, ChatGPT-4 performed comparable to physicians. Another notable finding was that submitting the identical query to ChatGPT-3.5 or ChatGPT-4 three different times had the same leading diagnosis only 60% of the time and overlap of all of the differential diagnoses only 70% of the time.

ChatGPT-4 output has also been found to align well with accepted guidelines for managing mild and severe depression, without showing the gender or socioeconomic biases sometimes observed among primary care physicians.(52) ChatGPT-4 has also been found to provide highly accurate (98.9%) medication recommendations as a second opinion in a dermatology treatment setting, but its reliability and comprehensiveness were deemed by researchers to need refinement for greater accuracy.(53)

Another study prompted ChatGPT-4 with symptoms of 194 diseases in the Mayo Clinic Symptom Checker, ChatGPT-4 achieved 78.8% accuracy in making the correct diagnosis.(54) The performance varied by clinical specialty, with best results for dentistry, endocrinology, and infectious diseases, and worst results for pulmonology, dermatology, and emergency medicine.

3.5 Assessing Clinical Reasoning

A number of studies have been based on instruments that assess clinical reasoning. In one study of clinical reasoning using 20 patient cases, GPT-4 was found to perform comparable to attending physicians and residents in diagnostic accuracy, correct clinical reasoning, and cannot-miss diagnosis inclusion.(55) In another pair of studies of diagnostic reasoning, physicians were assessed on a set of clinical vignettes developed to assess clinical decision support systems in the 1990.(56) Physicians were randomized for different vignettes to having conventional information resources with or without the addition of GPT-4 and assessed using an instrument to assess diagnostic reasoning.(57) Physicians using GPT-4 scored comparably to those not using it (76% vs. 73%), although those using ChatGPT-4 showed a trend to solving cases faster (565 vs. 519 seconds). GPT-4 alone scored much higher at 92%. In a similar randomized vignette study assessing clinical management decisions, physicians scored 6.5% higher using an LLM compared to those using conventional resources, but in this instance, GPT-4 alone did not do better.(58)

Another study of clinical reasoning was based on a framework developed to simulate a realistic clinical setting based on diagnostic accuracy, adherence to diagnostic and treatment guidelines, consistency in following instructions, ability to interpret laboratory results, and robustness to changes in instruction, information quantity and information order.(59) Based on a curated dataset based on the Medical Information Mart for Intensive Care (MIMIC) database of 2400 real patient cases and four common abdominal pathologies, it was found that LLMs did not accurately diagnose patients across all pathologies (performing significantly worse than physicians), follow neither diagnostic nor treatment guidelines, and were not able to interpret laboratory results. The authors also asserted that these LLMs could not be easily integrated into existing workflows because they often failed to follow instructions and were sensitive to both the quantity and order of information. One limitation of this study was the non-use of the known best-performing LLMs due to licensing restrictions that would not allow MIMIC data to be used in commercial versions of LLMs.

A Google LLM that has been studied with real users is the Articulate Medical Intelligence Explorer (AMIE). An initial study applied AIME in a randomized controlled trial that compared primary care physicians and the AMIE output that was judged by patient actors and specialist physicians.(60) The study was limited by the use of text-based dialogue for human-system interaction, but AMIE was found to outperform primary care physicians in history-taking, diagnostic accuracy, management reasoning, communication skills, and empathy. A second study developed a real-world dataset of 204 cardiology cases that included reports from electrocardiograms, echocardiograms, cardiac magnetic resonance imaging, genetic tests, and cardiopulmonary stress tests.(61) A ten-domain evaluation rubric was developed and used by cardiologists to evaluate the quality of diagnosis and clinical management plans for the cases entered into AMIE. The recommendations from AMIE were rated superior to general cardiologists for 5 of the 10 domains and equivalent for the others.

Another study of clinical problem-solving was a blinded observational comparative study conducted in the primary care setting in Sweden.(62) Responses from GPT-4 and real physicians to cases from a family medicine specialist examination were scored by blinded reviewers. In these complex primary care cases, GPT-4 and the follow-on GPT-4o performed worse than average human physicians. Recognizing the need to assess clinical reasoning with more complex dialogues and less with multiple-choice questions, the Conversational Reasoning Assessment Framework for Testing in Medicine or CRAFT-MD dataset was developed.(63) This dataset focuses on natural dialogues, using simulated agents to interact with LLMs in a controlled environment. The CRAFT-MD framework was found to show that LLMs performed notably worse in conversational settings compared to examination-based evaluations. The authors noted that more realistic testing approaches must be developed before LLMs can be safely integrated into clinical workflows and proposed a set of recommendations to align LLM evaluations with real-world clinical practice.

Although all of the studies presented so far have been based on textual data, some studies have assessed the use of foundational LLM models that include images and clinical tasks using them. In one study of image diagnosis cases using cases from the JAMA Clinical

Challenge and NEJM Image Challenge databases, OpenAI's GPT-4V demonstrated better image-interpretation accuracy than unimodal LLMs.(64) Another study based on NEJM Image Challenges found that GPT-4V performed comparatively to human physicians regarding multi-choice accuracy (81.6% vs. 77.8%).(65) GPT-4V also performed well in cases where physicians answered incorrectly, with about 78% accuracy. However, GPT-4V was also found to provide flawed rationales in cases where it made the correct final choices (35.5%), most prominently in image comprehension (27.2%).

An additional study combined PaLM with radiology reports along with an image encoder, and enabled detection of five findings on chest x-rays: atelectasis, cardiomegaly, consolidation, pleural effusion, and pulmonary edema.(66) Another study used a model for images and text that outperformed image-only and non-unified models for prediction of adverse events in pulmonary disease.(67) An additional study combined a locally-developed LLM and an image classifier trained on prior chest x-rays paired with reports.(68) The resulting system was assessed with new chest-rays in an emergency department and found to produce draft reports with clinical accuracy and textual quality that were comparable to on-site radiologist reports and provided higher textual quality than reports from off-site teleradiologists. Finally, another study used the LLaVA-Med LLM with image-report pairing to correctly answer questions from a visual question-answering dataset better than previous systems.(69)

One role for generative AI in radiology may be to assist radiologists in their reports. One study noted the ability of GPT-4 to identify missed diagnoses in preliminary reports of radiology trainees.(70) Another study found that GPT-4 detected errors in radiology reports (e.g., omissions, insertions, spelling, side confusion, and more) comparable to radiologists and did so much faster than them (average of 3.5 vs. 25 seconds).(71)

3.6 Summarization Tasks

As education often involves summarization of scientific articles, medical records, and other documents, LLMs have been assessed in summarization tasks. One study assessed GPT-4 at providing feedback to authors of computational biology scientific papers.(72) Assessing the PDFs of papers, the LLM feedback was found to have overlap comparable to between humans and was deemed higher for poorer-quality papers. Over half of authors of the papers found the generated feedback to be helpful or very helpful, and 82% found it more beneficial than feedback from at least some human reviewers. Another study looked at summaries of 140 evidence-based journal abstracts generated by ChatGPT-3.5 and found that the resulting summaries were 70% shorter than the mean abstract length and were found to have high quality, high accuracy, and low bias.(73) An additional system for synthesis of knowledge used a combination of gathering evidence, traversing citations, and synthesizing the results to perform better than humans in question-answering, summarization, and contradiction detection.(74)

ChatGPT-4 has also been assessed in the generation of lay summaries of scientific abstracts for a national clinical study recruitment platform, ResearchMatch.(75) ChatGPT-4 achieved 95.9% accuracy and 96.2% relevance across 192 summary sentences from 33 abstracts of clinical studies. A total of 85.3% of 34 volunteers rated ChatGPT-generated summaries as more accessible and 73.5% rated them more transparent than the original abstract. None of the summaries were rated as harmful by clinical experts.

3.7 Predictive Tasks

While this review distinguished at the onset a different between predictive and generative AI, the latter has been found in a number of instances to have predictive success as well. GPT-4 was shown to achieve accuracy of predicting cardiovascular disease comparable to the well-known Framingham predictive model with data from the UK Biobank and the Korean Genome and Epidemiology Study.(76) Other studies have assessed GPT-4 for predictive abilities in the emergency department (ED), finding suboptimal ability to predict acuity of patients in the ED(77) or predict admissions from ED to the hospital.(Glicksberg, 2024) However, better success has been found in identifying missed diagnoses in radiology residents' reports(70) and in increasing accuracy of rare disease diagnosis.(78)

3.8 Performance in Graduate Courses in Biomedicine and Health

Although the majority of biomedical and health applications of LLMs discussed so far have focused on medicine, some applications have focused on graduate-level study. One analysis compared LLM and student performance on graduate-level examinations in biomedical sciences.(79) The performance of GPT-4 on 9 examinations from courses in topics such as genomics, microbiology, and cellular and molecular biology was compared with actual student performance. GPT-4 was found to exceed the student average on 7 of 9 exams and for all student scores for 4 exams. GPT-4 performed very well on fill-in-the-blank, short-answer, and essay questions as well as questions on figures sourced from published manuscripts. It performed less well on questions with figures containing simulated data or those requiring hand-drawn answers. The analysis found that some model responses included hallucinations.

Another analysis compared student knowledge-assessment scores with prompting of 6 large-language model (LLM) systems as they would be used by typical students in a large online introductory course in biomedical and health informatics.(80) This course is taken by graduate, continuing education, and medical students. The state-of-the-art LLM systems were prompted to answer 10 MCQs each from the 10 units of the course and 33 final exam questions. Scores for 139 students (30 graduate students, 85 continuing education students, and 24 medical students) who took the course in 2023 were compared to the LLM systems. Google's Gemini scored highest across all assessments, but rest of the LLMs did well enough to achieve a passing grade for the course and scoring between the 50th and 75th percentiles of students (see Figure 3). The performance of the LLMs raised questions about student assessment in higher education, especially in courses that are knowledge-based and online.

Figure 3 – student scores for 25th, 50th (median), and 75th quartile of performance (thinner green, orange, and blue lines respectively) versus best-performing LLM, Gemini Pro (thicker black line) in the individual and aggregate unit assessments and the final examination for a large, online introductory biomedical and health informatics course.(80)

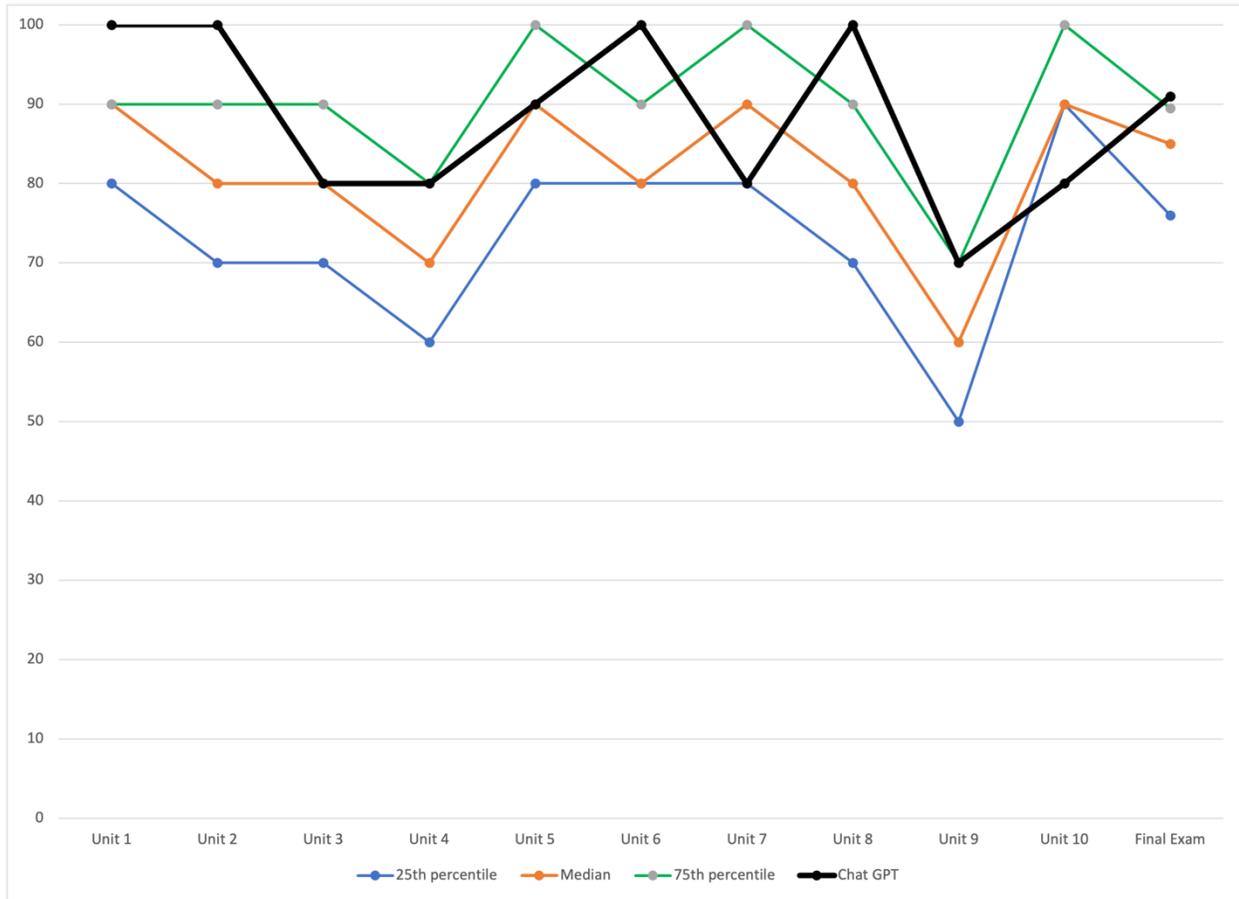

Another study from health informatics looked at the use of GitHub CoPilot in a programming course.(81) GitHub is a computer code repository system that has been integrated with an OpenAI LLM that aims to assist in the writing of computer code. The system was evaluated in a health informatics programming course for two types of programming problems, one for database queries in structured query language (SQL) and the other for computational tasks using the Python programming language. In general, the generated solutions worked well for simple and straightforward SQL and Python tasks but less well for more complex ones. It was also noted that some solutions were correct but did not take the most efficient approach to programming.

3.9 Performance in Academic Courses Beyond Biomedicine and Health

Generative AI systems have also been assessed in education beyond biomedicine and health. One report describing the release of a new open-source LLM, Llama 3, noted that like other major LLMs, it was capable of passing high school advanced placement and college prep tests.(82) At the college undergraduate level, examinations for five courses in psychology at a university in the United Kingdom were written by GPT-4 and achieved grades higher than the average of students taking the exams in the courses.(83) It was also found that 94% of the AI submissions went undetected. Success has been achieved at even more advanced academic levels, such as on the so-called Graduate-Level Google-Proof Q&A Benchmark (GPQA), a challenging set of questions from biology, chemistry, and physics.(84) The OpenAI o1 LLM was found to exceed the scores of PhD-level students on this dataset.(85)

One area where generative AI has had major impact has been computer science. In an overview, it was noted that LLMs are very capable computer programmers, raising questions about the teaching and even the practice of computer programming going forward.(86) This author noted that LLMs are capable of generating solutions to problems typical of introductory programming courses, raising concerns around potential student overreliance and misuse. A comprehensive analysis of many different Python programming tasks found that GPT-4 could create computer code even for complex tasks but still required humans to ensure validity and accuracy of the code.(87) Another author noted that computer science educators at all types of institutions must monitor changes and adjust curriculum accordingly.(88)

The success of LLMs in academic tasks has been demonstrated in other disciplines. For law students, GPT-4 was found to modestly improve student performance in a number of legal tasks and complete it markedly faster.(89) Likewise, LLMs have been found to be able to create functional pipelines in data science tasks.(90,91) Another study found that GPT-4 was able to solve novel and difficult tasks that span mathematics, coding, vision, medicine, law, psychology, and more, without the need for any special prompting.(92)

Some studies have looked at use of LLMs outside the academic arena, but their mixed results have implications for education. One study looked at a creative divergent thinking task called the Alternative Uses Task and found that ChatGPT-3.5 and 4 outperformed average but not the best humans.(93) Another study found that ChatGPT-4 exceeded other LLMs at a variety of general human language tasks but performed less well on reasoning tasks and was still prone to hallucinations.(94) An additional study that assigned writing tasks to 453 college-educated professionals found a 40% decrease in time and an 18% improvement in quality for the half of subjects assigned to use ChatGPT.(95) In business, GPT-4 has been found to outperform analysts in ability to predict earnings changes in companies.(96,97) In software engineering, 3 randomized controlled trials showed an average of 26% productivity gains, with the gains higher for workers with less experience.(98)

Another study at a global management consulting firm randomized consultants to using ChatGPT-4 or not in their work.(99) Those assigned to ChatGPT-4 were found to be more productive, completing an average of 12.2% more tasks, and more quickly completing tasks 25.1% of the time. Consultants assigned ChatGPT-4 were also found to produce higher-quality results, measured to be more than 40% higher quality compared to control group. For some tasks, however, such as combining qualitative and quantitative data, ChatGPT-4 performed less well, leading the authors to note there was a so-called jagged technological frontier, where some tasks were easily done by AI but others not.

4. Limitations of Generative AI

The limitations of generative AI are well-known, and will be reviewed here from the perspective of education. From the earliest days of ChatGPT, it was seen that LLMs sometimes "hallucinate." Indeed, the Dictionary.com 2023 Word of the Year was hallucinate, mostly based on its connection to ChatGPT and other generative AI.(100) This section will describe the following limitations of generative AI:
1. Factuality of LLMs
2. Citations and Attribution
3. Data Bias, Leakage, and Drift
4. Detecting Use of Generative AI
5. Real-World and Safe Use

4.1 Factuality of LLMs

Going beyond hallucinations and confabulations, Augenstein has described "factuality challenges" for LLMs, which include:(101)
- Undersourcing, i.e., lacking credible references to statements made
- Truthfulness, i.e., their hallucinations
- Speak with a confident tone and fluent style that may mislead users who are less familiar with the facts about a given topic
- Allow direct use by users and are easy to access
- Halo effect from being knowledgeable in other domains than the one of current interest to a user
- Public perception that LLMs are all-knowing
- Unreliable evaluation, leading to excess positivity about their functionality

4.2 Citations and Attribution

A related concern to factuality is the tendency for LLMs to generate fabrications and errors in citations, an important aspect of scientific and scholarly communication. One study prompted ChatGPT to produce short literature reviews on 42 multidisciplinary topics. Analysis found that 55% of GPT-3.5 citations and 18% of GPT-4 citations were fabricated. In addition, 43% of real GPT-3.5 citations and 24% of real GPT-4 citations included substantive errors.(102)

A related concern is that current LLMs are not necessarily good at attributing their information with references. One study looked at reference validity and attribution for several LLMs.(103) It was found that GPT-4 as used in Microsoft CoPilot, had very high validity of URL sources cited, with other LLMs performing more poorly. But even GPT-4 in CoPilot only provided statement-level support for 70% of its assertions and response-level support for 54%, with the others faring worse. Even CoPilot failed to cite any sources for around 20% of prompts, with others having higher rates of non-citation. Another issue that this study raised was attribution sources behind paywalls that users might not be able to access.

These results led one author to note that from the perspective of biomedicine and health, especially those in academic use of searching, may have concerns for authoritativeness, timeliness, and contextualization of search results. In other words, we often search not just to find an answer, but also to find out where the answer came from, who wrote it, and what methods they used.(104)

4.3 Data Bias, Leakage, and Drift

LLMs have also been documented to perpetuate bias that is present in many information sources that are used for training LLMs, including scientific literature and other information on the Internet. One study assessed bias by asking eight clinical questions of four different LLMs, such as estimated glomerular filtration rate or eGFR, lung capacity, and pain threshold.(105) The four LLMs – ChatGPT-3.5, Bard, Claude, and GPT-4 – all were found to recapitulate what the authors called harmful, race-based medicine.

Another study analyzed standardized clinical vignettes from the publication NEJM Healer.(106) GPT-4 was found to be more likely to include diagnoses that stereotyped certain races, ethnicities, and genders, and did not model appropriate demographic diversity of various medical conditions in the vignettes. Outside of medicine, one study assessed AI-generated content (AIGC) produced by new articles in the New York Times and Reuters and found substantial gender and racial biases.(107) AIGC generated by each LLM exhibited notable discrimination against female and black individuals. Among the LLMs, AIGC generated by ChatGPT demonstrated the lowest level of bias, and ChatGPT was the only LLM able to decline content generation when biased prompts were entered.

Additional concerns for LLMs include issues around the data used to train them. One issue is data leakage, which results when there is contamination of model input features with outcome information. Or put another way, when training data makes it into the test data and potentially influences the result and may artificially improve the metric being used to evaluate the model. One analysis found at least 294 papers across 17 different scientific disciplines and identified eight reasons for data leakage. Whatever the reason, the usual cause was training data making its way into or somehow otherwise influencing test data, potentially biasing the performance results from the model.(108)

A related concern is dataset shifts due to changes that may occur in technology, population and setting, and clinician or patient behavior.(109) Diseases may change over time as well, with a good example COVID-19, which is a different disease from when the pandemic began in 2020.(110) Another type of information that changes over time is genetic information, which may be due to new associations found in research or new technologies used. One study developed a chained, two-prompt GPT-4 sequence from a training set of 45 article-variant pairs for the automated classification of functional genetic evidence in the scientific literature. The first prompt asked GPT-4 to supply functional evidence in a given article for the variant of interest or indicate the lack of such evidence. For articles in which GPT-4 designated functional evidence, a second prompt asked it to classify the evidence into categories of pathogenic, benign, or intermediate/inconclusive. A test set of 72 manually classified article-variant pairs found substantial variability over 2.5-months in the results both within prompts entered in rapid succession or across days.(111)

4.4 Detecting Use of Generative AI

One desired feature of LLMs in many applications, including those from education, would be the ability for people or computer algorithms to be able to detect their use. Certainly in education, it would be valuable to be able to discern the use of LLMs, especially in situations where educators might not want them to be used. Unfortunately, research shows that both machine and human detection of LLM usage is inconsistent at best.(112)

One system focused on research abstracts from scientific journals was shown to have a 98% rate of detection accuracy and perform much better than humans.(113) Another pair of studies found a machine learning model could distinguish scientific writing from ChatGPT,(114) including its focused use in chemistry journals.(115) However, another study found that light paraphrasing undermined generative AI detectors (Sadasivan, 2023) A further evaluation of 11 Web-based detectors found simple modifications undermined detectors, such as introduction of minor grammatical errors and substitution of Latin with similar Cyrillic letters.(116) Another study found that the already-low accuracy rates (39.5%) of LLM detectors showed further reductions in accuracy (17.4%) when faced with manipulated content, with some techniques proving more effective than others in evading detection.(117) Another concern for LLM detectors is that some systems are more likely to classify non-native English writing as AI-generated.(118)

Humans are not good detectors of LLM writing either. One study in a consulting firm found that humans were not able to discern AI writing well.(99) Another study looked at so-called compelling disinformation, noting that humans were unable to distinguish between true and false tweets generated by GPT-3 or written by real Twitter users.(119) An additional study found that peer-reviewers were not able to distinguish AI-generated from human-generated text in the journal article peer review process for an applied linguistics

journal.(120) Finally, randomized-controlled experiments investigating novice and experienced teachers' ability to identify AI-generated texts showed that generative AI can write student essays writing in ways that were undetectable by teachers. Teachers were found to be overconfident in their source identification, and AI-generated essays were assessed more positively than student-written texts.(121)

4.5 Real-World and Safe Use

Another broad concern is that most of the studies in the previous section showing impressive application of LLMs were conducted in simulated settings that may not reflect real-world use of generative AI. Indeed, there are few clinical trials that have assessed patient or health care delivery outcomes using all types of AI.(122) There are also few assessments of the safety of generative AI, with some noting that legal liability for clinicians still rests with practitioners themselves when they use AI tools.(123)

5. Impact of Generative AI on Education

We have already seen that impact of generative AI on activities associated with education is profound and varied. This section of the review will describe recommendations for best practices going forward. Many thought leaders have expressed bold opinions for the use of AI in medical (and really all health professions) education. The Dean of Harvard Medical School has noted, "AI should and will change medical school."(124) Clinicians must be prepared to practice in a world of AI.(125) Medical schools face dual challenges of needing to teach about AI in practice but also adapt to its use by learners and faculty.(126) Physicians must be prepared for the "clinical algorithm era."(127) Others conjecture that LLMs will change education and not destroy it, noting that assessment may already be broken and stating, "if ChatGPT makes it easy to cheat on an assignment, teachers should throw out the assignment rather than ban the chatbot."(128) Another writer has asked, "Are we just grading robots?"(129) This section will cover the following topics:
1. Competencies for Use of AI
2. Prompt Engineering
3. Considerations for Learning
4. Role in Clinical Education
5. Roles Outside Clinical Education

5.1 Competencies for Use of AI

Some work has focused on student competencies for AI, mainly focused on use by students who will become practicing clinicians. One well-known set of competencies in clinical informatics was developed a decade ago(130) and was recently expanded to include AI:(131)
- Find, search, and apply knowledge-based information to patient care and other clinical tasks

- Effectively read from, and write to, the electronic health record for patient care and other clinical activities
- Use and guide implementation of clinical decision support (CDS)
- Provide care using population health management approaches
- Protect patient privacy and security
- Use information technology to improve patient safety
- Engage in quality measurement selection and improvement
- Use health information exchange (HIE) to identify and access patient information across clinical settings
- Engage patients to improve their health and care delivery though personal health records and patient portals
- Maintain professionalism through use of information technology tools
- Provide clinical care via telemedicine and refer patients as indicated
- Apply personalized/precision medicine
- Participate in practice-based clinical and translational research
- Use and critique AI applications in clinical care

Others competency frameworks have been proposed that are more specific to AI. One focuses on the use of AI in primary care, proposing competencies in 6 domains:(132)
- Foundational knowledge – what is this tool?
- Critical appraisal – should I use this tool?
- Medical decision making – when should I use this tool?
- Technical use – how do I use this tool?
- Patient communication – how should I communicate with patients regarding the use of the tool?
- Unintended consequences (cross-cutting) – what are the "side effects" of this tool?

Another framework, also with 6 areas, focuses on the use of AI-based tools by healthcare professionals more broadly:(133)
- Basic knowledge of AI – explain what AI is and describe its healthcare applications
- Social and ethical implications of AI – explain how social, economic, and political systems influence AI-based tools and how these relationships impact justice, equity, and ethics
- AI-enhanced clinical encounters – carry out AI-enhanced clinical encounters that integrate diverse sources of information in creating patient-centered care plans
- Evidence-based evaluation of AI-based tools – evaluate the quality, accuracy, safety, contextual appropriateness, and biases of AI-based tools and their underlying datasets in providing care to patients and populations
- Workflow analysis for AI-based tools – analyze and adapt to changes in teams, roles, responsibilities, and workflows resulting from implementation of AI-based tools

- Practice-based learning and improvement regarding AI-based tools – participate in continuing professional development and practice-based improvement activities related to use of AI tools in healthcare

An additional framework presents a matrix of competencies for two types of clinicians, general and AI-specialist, in three broad domains of health informatics, AI, and generative AI/LLMs.(134)

5.2 Prompt Engineering

Another important competency in the era of generative AI is prompting, which is sometimes called prompt engineering. One primer focused on prompt engineering in the biomedical and health context and recommends:(135)
- Be as specific as possible
- Describe the setting and provide the context around the question
- Experiment with different prompt styles
- Identify the overall goal of the prompt first
- Ask the LLM to play roles
- Iterate and refine prompts
- Use threads of prompts
- Ask open-ended questions
- Request examples
- Provide temporal awareness
- Set realistic expectations
- Use the one- or few-shot prompts
- Prompting LLMs for prompts

Another author notes a number of considerations for healthcare prompt engineering, raising issues that should be considered in the prompting process.(136) These include recommendations similar to those above, but also some ethical and legal issues, such as not entering any personally identifiable information that may become part of the LLM; compliance with privacy standards, such as the Health Insurance Portability and Accountability Act (HIPAA) and the General Data Protection Regulation (GDPR), and aligning use with medical ethics.

5.3 Considerations for Learning

One concern for generative AI is the ease by which it can answer questions, pass tests, and substitute for deeper learning and understanding a topic, all of which potentially undermining the ability of students to engage in critical thinking, both in their education and into their real-world professional work that follows their education. Although educators often try to make learning an enjoyable process, a recent meta-analysis found an association between mental effort and aversive affect, i.e., learning takes effort that is not always pleasant.(137) This can lead to learners taking shortcuts, a process that started with use of Web search engines and now includes the use of generative AI. One recent study assessed 2433 over an 11-year period in 12 different college lecture courses.(138) The result found that the percent of students who did not show performance benefit by correctly answering homework questions increased from 14% in 2008 to 55% in 2017, presumably due to taking shortcuts to answering homework questions and not learning the underlying material as well. In the last 2 years of the study, when students were asked how they did their homework, students who benefitted from homework reported generating their own answers while students who reported copying the answers from another source did not benefit from homework.

Other studies have specifically looked at the impact of ChatGPT in learning. One study assessed learning Python programming in a data science course and found that the use of LLMs as personal tutors by asking for explanations improved learning outcomes but that excessively asking LLMs to generate solutions impaired learning.(139) The latter was made more adverse by allowing copy-and-paste availability. Another study evaluated learning math at the high school level.(140) Both a plain version of ChatGPT-4 and one augmented with prompts aiming to provide more of a tutor mode were found to increase performance on exams by 48% and 127% respectively. However, when the LLMs were taken away from students, subsequent scores averaged 17% worse than baseline for those using the plain version and a return to baseline for the tutor version. These studies show that LLMs may have adverse consequences on student learning, allowing easy answering of questions but not resulting in mastery of the material.

5.4 Role in Clinical Education

A number of authors have written about the role of generative AI in clinical education specifically. One narrative review looked at the potential of LLMs for medical education, noting potential advantages for students and faculty but also challenges.(141) This narrative review noted advantages to students, such as more direct access to information, personalizing learning activities, and facilitating development of clinical skills. Potential benefits for faculty and instructors include the development of innovative approaches to pedagogy for complex medical concepts and facilitating student engagement. However, the review also noted a number of challenges, such as leading to academic misconduct, becoming over-reliant on AI, diluting critical thinking skills, concerns for the veracity and reliability of LLM-generated content, and implications for all of these on teaching staff.

Another paper proposed a set of recommendations for medical faculty and their institutions.(142) The authors advocated that educators must increase their knowledge of AI, understand the current landscape for its use in medical practice and education, review strategies for successful AI integration into education, and become stewards of its ethical use. Likewise, institutions should review and revise school policies, creating new policies as needed, regarding use of generative AI; support faculty development about AI and provide resources for teaching, and offering information-checking tools for originality and plagiarism to faculty.

An additional paper noted a number of use cases for how LLMs can be integrated into medical education:(143)
- Practice generating differential diagnoses
- Streamlining the wide array of study resources to assist with devising a study plan
- Serving as a simulated patient or medical professor for interactive clinical cases
- Helping students review multiple-choice questions or generating new questions for additional practice
- Digesting lecture outlines and generating materials for flash cards
- Organizing information into tables to help build scaffolding for students to connect new information to previous knowledge

Other uses for LLMs in health professions education that have been forth include creating checklists for common presentations and generate templates for common clinical scenarios,(144) presenting potential patient problems to nurses and guiding students through clinical processes,(145) and generating radiology board-style MCQs.(146)

5.5 Roles Outside Clinical Education

Additional recommendations for use of generative in AI for teaching have come from outside biomedicine and health. A couple of business school professors have written about "assigning AI," noting the different ways it might be used:(147)
- Mentor – providing feedback
- Tutor – direct instruction
- Coach – prompt metacognition
- Teammate – increase team performance
- Student – receive explanations
- Simulator – deliberate practice
- Tool - accomplish tasks

One of these authors has written a book about AI-human "co-intelligence"(148) and posted a library of prompts for use by educators.(149) The two have also promoted the notion of instructors as innovators enabling novel forms of practice and application including simulations, mentoring, coaching, and co-creation.(150) A number of other authors have written books describing approaches to using AI in education. One of these books covers topics such as policies and cheating, feedback and roleplaying, assignments, writing, and assessments.(151) An additional business school professor described his approach to using LLMs to teach management sciences, noting that they should be used in writing assignments in an iterative manner for the duration of a course and that multiple-choice, brief response, and fill-in-the-blank tests should be avoided.(152)

Clearly education at all levels, especially higher education, will need to adapt to AI. Another author notes that colleges and universities will need to move from "foundational to multifaceted" AI education, leverage experiential learning to use AI as well as surmount it, and use lifelong learning for reinvention, not just to acquire new skills.(153)

6. Conclusions

Generative AI is having profound impact on biomedical and health professions and their education. It has been rapidly taken up by clinicians, students, and indeed all of society. Generative AI has been found to perform as good as human experts on many but not all intellectual tasks. In biomedicine and health, this includes medical board exams, answering clinical questions, solving clinical cases and applying clinical reasoning, summarizing information, and performing well in academic courses. For many activities, it helps less experienced individuals more than experts, although in other situations it can lead all people astray. There are also a number of limitations of generative AI, such as answers including hallucinations and confabulations, inability to cite references and provide attribution, providing answers based on biased data, and algorithms and humans being poor at discerning text from generative AI. In addition, many of the studies have been done in simulated settings and not real-world practice. Generative AI is likewise being used widely in education, and a number of authors have provided recommendations for its optimal use in pedagogy. Clearly there are challenges for its use, especially if it undermines the acquisition of knowledge and skills for professional activity. However, all students and faculty, in biomedicine and health and beyond, must have a thorough understanding of it and be competent in its use.

This author has witnessed a number of transformations in education, including the transition from slide rules to calculators while in high school in the 1970s and the emergence of Google and other Internet search engines as young faculty in the 1990s. The transition to generative AI, however, is more probably more profound, since the earlier transitions simplified the steps of problem solving and critical thinking whereas generative AI potentially replaces them. Educators must develop new approaches to teaching and student assessment in the era of generative AI, while healthcare, informatics, and educational professionals must be competent with AI as much as any other tool in professional practice.

7. Future Issues

Future issues for generative AI in education can be grouped into general issues for professional practice with its use and also concerns specific to education. Future issues for LLMs in biomedical and health professional practice include:
- Will performance of LLMs continue to improve in biomedical and health tasks, or will it level off?

- How will the use of LLMs be optimized and validated beyond simulated use and implemented in the larger workflow of biomedical and health professional practice that includes clinical care, research, administration, and other areas?
- What evidence will be required of assertions students and professionals obtain from generative AI?

More specific future issues for education include:
- What will be the optimal policies for use of LLMs in education?
- How will we assess student learning when generative AI tools are readily available?
- Knowing that in most biomedical and health disciplines, students can no longer memorize all the knowledge and skills of their disciplines, so what core must be mastered to provide a foundation and perspective that will enable advanced thinking?
- How will students and instructors minimize overreliance on LLMs that may undermine their professional broader professional competence?

10. Xiao H, Zhou F, Liu X, Liu T, Li Z, Liu X, et al. A comprehensive survey of large language models and multimodal large language models in medicine. Information Fusion. 2024 Dec 23;102888.
11. Nam J. BestColleges.com. 2023 [cited 2023 Dec 13]. 56% of College Students Have Used AI on Assignments or Exams | BestColleges. Available from: https://www.bestcolleges.com/research/most-college-students-have-used-ai-survey/
12. AI Chatbots in Schools - Findings from a Poll of K-12 Teachers, Students, Parents, and College Undergraduates [Internet]. Impact Research; 2024 May. Available from: https://8ce82b94a8c4fdc3ea6d-b1d233e3bc3cb10858bea65ff05e18f2.ssl.cf2.rackcdn.com/bf/24/cd3646584af89e7c668c7705a006/deck-impact-analysis-national-schools-tech-tracker-may-2024-1.pdf
13. Bick A, Blandin A, Deming DJ. The Rapid Adoption of Generative AI [Internet]. National Bureau of Economic Research; 2024 [cited 2024 Oct 29]. (Working Paper Series). Available from: https://www.nber.org/papers/w32966
14. Blease CR, Locher C, Gaab J, Hägglund M, Mandl KD. Generative artificial intelligence in primary care: an online survey of UK general practitioners. BMJ Health Care Inform. 2024 Sep 17;31(1):e101102.
15. Presiado M, Montero A, Lopes L, Published LH. KFF Health Misinformation Tracking Poll: Artificial Intelligence and Health Information [Internet]. KFF. 2024 [cited 2024 Oct 18]. Available from: https://www.kff.org/health-misinformation-and-trust/poll-finding/kff-health-misinformation-tracking-poll-artificial-intelligence-and-health-information/
16. Jin D, Pan E, Oufattole N, Weng WH, Fang H, Szolovits P. What Disease Does This Patient Have? A Large-Scale Open Domain Question Answering Dataset from Medical Exams. Applied Sciences. 2021 Jan;11(14):6421.
17. Kung TH, Cheatham M, Medenilla A, Sillos C, De Leon L, Elepaño C, et al. Performance of ChatGPT on USMLE: Potential for AI-assisted medical education using large language models. PLOS Digit Health. 2023 Feb;2(2):e0000198.
18. Nori H, King N, McKinney SM, Carignan D, Horvitz E. Capabilities of GPT-4 on Medical Challenge Problems [Internet]. arXiv; 2023 [cited 2023 Jul 17]. Available from: http://arxiv.org/abs/2303.13375
19. Corrado G, Barral J. Google Research. 2024 [cited 2024 Oct 2]. Advancing medical AI with Med-Gemini. Available from: http://research.google/blog/advancing-medical-ai-with-med-gemini/
20. Horvitz E, Nori H, Usuyama N. Run-time strategies in foundation models: from Medprompt to OpenAI o1-preview [Internet]. Microsoft Research. 2024 [cited 2024 Dec 2]. Available from: https://www.microsoft.com/en-us/research/blog/advances-in-run-time-strategies-for-next-generation-foundation-models/
21. Nori H, Usuyama N, King N, McKinney SM, Fernandes X, Zhang S, et al. From Medprompt to o1: Exploration of Run-Time Strategies for Medical Challenge Problems and Beyond [Internet]. arXiv; 2024 [cited 2024 Dec 1]. Available from: http://arxiv.org/abs/2411.03590
25

37. Goodman RS, Patrinely JR, Stone CA, Zimmerman E, Donald RR, Chang SS, et al. Accuracy and Reliability of Chatbot Responses to Physician Questions. JAMA Netw Open. 2023 Oct 2;6(10):e2336483.
38. Duong D, Solomon BD. Analysis of large-language model versus human performance for genetics questions. Eur J Hum Genet. 2023 May 29;
39. Dash D, Thapa R, Banda JM, Swaminathan A, Cheatham M, Kashyap M, et al. Evaluation of GPT-3.5 and GPT-4 for supporting real-world information needs in healthcare delivery [Internet]. arXiv; 2023 [cited 2023 Dec 21]. Available from: http://arxiv.org/abs/2304.13714
40. Rodman A, Buckley TA, Manrai AK, Morgan DJ. Artificial Intelligence vs Clinician Performance in Estimating Probabilities of Diagnoses Before and After Testing. JAMA Netw Open. 2023 Dec 1;6(12):e2347075.
41. Zakka C, Shad R, Chaurasia A, Dalal AR, Kim JL, Moor M, et al. Almanac — Retrieval-Augmented Language Models for Clinical Medicine. NEJM AI. 2024 Jan 25;1(2):AIoa2300068.
42. Low YS, Jackson ML, Hyde RJ, Brown RE, Sanghavi NM, Baldwin JD, et al. Answering real-world clinical questions using large language model based systems [Internet]. arXiv; 2024 [cited 2024 Oct 29]. Available from: http://arxiv.org/abs/2407.00541
43. Xie Y, Wu J, Tu H, Yang S, Zhao B, Zong Y, et al. A Preliminary Study of o1 in Medicine: Are We Closer to an AI Doctor? [Internet]. arXiv; 2024 [cited 2024 Oct 7]. Available from: http://arxiv.org/abs/2409.15277
44. Semigran HL, Levine DM, Nundy S, Mehrotra A. Comparison of Physician and Computer Diagnostic Accuracy. JAMA Intern Med. 2016 Dec 1;176(12):1860–1.
45. Benoit JRA. ChatGPT for Clinical Vignette Generation, Revision, and Evaluation [Internet]. medRxiv; 2023 [cited 2023 Aug 30]. p. 2023.02.04.23285478. Available from: https://www.medrxiv.org/content/10.1101/2023.02.04.23285478v1
46. Levine DM, Tuwani R, Kompa B, Varma A, Finlayson SG, Mehrotra A, et al. The diagnostic and triage accuracy of the GPT-3 artificial intelligence model: an observational study. Lancet Digit Health. 2024 Aug;6(8):e555–61.
47. Kanjee Z, Crowe B, Rodman A. Accuracy of a Generative Artificial Intelligence Model in a Complex Diagnostic Challenge. JAMA. 2023 Jul 3;330(1):78–80.
48. Eriksen AV, Möller S, Ryg J. Use of GPT-4 to Diagnose Complex Clinical Cases. NEJM AI [Internet]. 2023 Nov 9; Available from: https://onepub-media.nejmgroup-production.org/ai/media/ec2de32e-9aa9-49f0-8f37-45becf6be3ed.pdf
49. McDuff D, Schaekermann M, Tu T, Palepu A, Wang A, Garrison J, et al. Towards Accurate Differential Diagnosis with Large Language Models [Internet]. arXiv; 2023 [cited 2023 Dec 5]. Available from: http://arxiv.org/abs/2312.00164
50. Rao A, Pang M, Kim J, Kamineni M, Lie W, Prasad AK, et al. Assessing the Utility of ChatGPT Throughout the Entire Clinical Workflow: Development and Usability Study. J Med Internet Res. 2023 Aug 22;25:e48659.
51. ten Berg H, van Bakel B, van de Wouw L, Jie KE, Schipper A, Jansen H, et al. ChatGPT and Generating a Differential Diagnosis Early in an Emergency Department Presentation. Ann Emerg Med. 2024 Jan;83(1):83–6.